# Agentic AI for Financial Crime Compliance

Henrik Axelsen
Copenhagen Business School
ha.digi@cbs.dk

Valdemar Licht
University of Copenhagen
joli@di.ku.dk

Jan Damsgaard
Copenhagen Businss School
jd.dig@cbs.dk

## Abstract

*The cost and complexity of financial crime compliance (FCC) continue to rise, often without measurable improvements in effectiveness. While AI offers potential, most solutions remain opaque and poorly aligned with regulatory expectations. This paper presents the design and deployment of an agentic AI system for FCC in digitally native financial platforms. Developed through an Action Design Research (ADR) process with a fintech firm and regulatory stakeholders, the system automates onboarding, monitoring, investigation, and reporting, emphasizing explainability, traceability, and compliance-by-design. Using artifact-centric modeling, it assigns clearly bounded roles to autonomous agents and enables task-specific model routing and audit logging. The contribution includes a reference architecture, a real-world prototype, and insights into how Agentic AI can reconfigure FCC workflows under regulatory constraints. Our findings extend IS literature on AI-enabled compliance by demonstrating how automation, when embedded within accountable governance structures, can support transparency and institutional trust in high-stakes, regulated environments.*

## 1. Introduction

Anti-Money Laundering (AML) refers to the policies, laws, and procedures used to prevent, detect, and report money laundering, terrorist financing (CFT), and transactions involving sanctioned entities (Financial Action Task Force (FATF), 2025). The broader term, Financial Crime Compliance (FCC), also includes fraud, bribery, corruption, market abuse, and tax evasion. In practice, institutions treat AML and FCC as overlapping domains, often handled within a unified compliance architecture. We adopt this convention, using the terms interchangeably to reflect the integrated nature of analytics, monitoring, and reporting systems.

The cost of FCC is rapidly escalating. In 2023 alone, global FCC expenditures exceeded USD 200 billion, roughly 5–10% of the estimated USD 3.1 trillion in illicit financial flows, or ~3% of global GDP (Forrester, 2023; LexisNexis, 2024). While these costs average USD 25 per person globally, they rise significantly in mature markets: USD 270 in Poland, USD 670 in the Netherlands, and nearly USD 700 in the UK, reflecting both regulatory pressure and laundering risk (LexisNexis, 2024). The UK and its overseas territories are implicated in an estimated 40% of global laundering flows (AML Intelligence, 2024), while the U.S. accounts for approximately 10% (Sanction Scanner, 2025).

Costs are not evenly distributed across FCC functions. Approximately two-thirds of the burden lies in customer due diligence (CDD) and know-your-customer (KYC) processes – identity verification, screening, and monitoring – while transaction monitoring and case management account for most of the rest (LexisNexis, 2024). Figure 1 illustrates the 8-step FCC process.

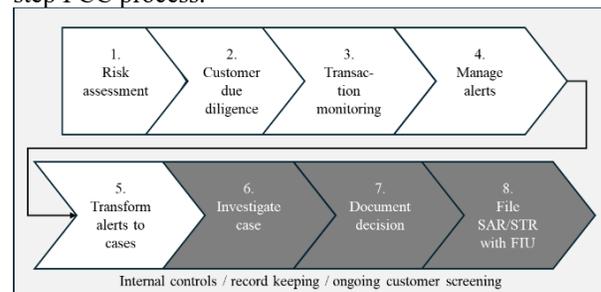

**Figure 1. The FCC process.**

The combined process represents a major operational cost centre for regulated entities. Regulatory benchmarks suggest an average suspicious transaction / activity report (STR/SAR) case requires ~2 hours, with complex cases up to 21 hours (Financial Crimes Enforcement Network, 2020; RegTech Consulting, 2020). These estimates understate total costs, as they only cover the final process steps 6-8 in Figure 1, which, as mentioned, account for roughly one third of total effort.

Despite rising costs, the effectiveness of FCC remains questionable. In Denmark, as an example of a digitally advanced jurisdiction, over 80,000 SARs/STRs were submitted in 2023, but fewer than half received follow-up, and fewer than 15% of obligated entities had integrated with the national STR platform (Hvidvasksekretariatet, 2023). This imbalance reflects a systemic tension: regulatory pressure and/or the absence of clear definitions can lead to misaligned reporting expectations, often also outpacing organizations' ability to integrate insights and take meaningful action.

Enforcement outcomes are similarly limited, with global seizure rates estimated below 1% (Pol, 2020), while laundering costs for criminal organizations remain competitive at just 3% in loss or fees (AP News, 2024). Recent policy research reinforces this problem: despite massive investments in compliance infrastructure, there is little evidence that money laundering has become more costly or difficult, and the system's main output remains intelligence rather than prevention (Nazzari & Reuter, 2025). This underscores the need for design approaches that can reduce inefficiencies and realign compliance with actual regulatory outcomes.

Traditional AML systems – often based on static rule engines or narrow machine learning (ML) models – struggle with high false positives, explainability limitations, and poor scalability. These limitations hinder proactive risk detection and significantly increase manual workload. The emergence of agentic artificial intelligence (AI) systems offers a new architectural approach. In contrast to isolated AI models or simple generative AI (GenAI) assistants, agentic AI systems combine autonomous decision agents with orchestrated workflows, enabling explainable, goal-directed, and auditable actions across multiple tasks (Acharya et al., 2025; Mukherjee & Chang, 2025; Sapkota et al., 2025).

In this study, we focus on the design of an agentic AI system for FCC in digitally native financial platforms. Agentic AI refers to AI systems composed of autonomous agents that can pursue sub-goals within regulatory workflows, using both large language models (LLMs) and structured logic, while being governed by guardrails that enforce compliance principles. These systems differ from generic automation in that they embed institutional roles, lifecycle logic, and escalation paths directly into agent behavior. When applied to FCC, such systems enable a shift toward what we may term 'agentic compliance' – a design paradigm where regulatory rules are enacted by AI agents with structured oversight and traceable accountability, rather than as static rules or manual processes managed by humans.

The target setting of our prototype – a digitally native platform – refers to a new category of financial services that operate natively in tokenized environments, such as blockchain-based gaming or decentralized finance. These platforms present novel compliance challenges: high transaction volumes, pseudonymity, and jurisdictional fragmentation (Denise et al., 2023). Our implementation involves a fintech startup located in Denmark and undergoing licensing as a Crypto-Asset Service Provider (CASP) under the EU's Markets in Crypto-Assets (MiCA) regulation, with non-fungible token (NFT) based micro-transactional flows.

Within this context, we designed and deployed a prototype agentic AI system to automate the core FCC lifecycle steps: onboarding, transaction monitoring, alert triage, case investigation, and SAR/STR reporting. The system integrates LLMs with rule-based logic, structured workflows, and explainability protocols.

We pose the following research question: *How can agentic AI systems be designed to support scalable, explainable, and regulation-aligned FCC in digitally native financial platforms?*

By combining agentic orchestration, compliance-by-design principles (Lohmann, 2013), and embedded explainability, we propose a practical framework that extends Information Systems (IS) literature on AI-enabled compliance.

## 2. Related work

Research on financial regulation and compliance has long been central to Information Systems (IS) scholarship. The emergence of RegTech captures how digital technologies reshape compliance processes through automation, auditability, and data-driven risk management (Gozman et al., 2018). Foundational IS work has emphasized how information systems mediate the interpretation of regulatory rules, create accountability structures, and support compliance-by-design approaches (Lohmann, 2013). These perspectives highlight compliance as not merely technical but socio-technical: an interplay of rules, organizational routines, and infrastructures.

*AI in Financial Crime Compliance*: Over the past decade, the rise of artificial intelligence (AI) and machine learning (ML) has fueled attempts to enhance anomaly detection, fraud prevention, and transaction monitoring. Prior studies show ML methods can reduce false positives and improve detection accuracy (Javaid, 2024; Korkanti, 2024). However, such systems often rely on black-box models with limited explainability, creating friction with regulators and compliance officers (Kute et al., 2021). This tension has given rise to explainable AI (XAI) approaches that embed interpretability into compliance workflows, aiming to build trust in AI-mediated decision-making. While XAI techniques provide transparency, their integration into operational compliance systems remains limited.

*Compliance-by-Design and Institutional Logic*: A recurring theme in IS literature is embedding compliance requirements into system architectures from the outset, an approach captured by compliance-by-design (Lohmann, 2013). This principle ensures regulatory obligations are not retrofitted but structurally enforced in workflows and system logic. Complementary perspectives from complexity-oriented IS research (Benbya et al., 2020) and innovation networks (Lyytinen et al., 2016) conceptualize compliance as dynamic, shaped by evolving data flows, actors, and institutional logics. Such views reinforce the need to design compliance systems that balance automation with governance and adaptability.

*RegTech and Emerging Digital Platforms*: RegTech scholarship has examined how information systems transform compliance through automation and digital infrastructures (Gerlings & Constantiou, 2023). Yet most studies focus on incremental improvements in transaction monitoring or fraud detection. Emerging digitally native platforms such as NFT marketplaces and permissionless public blockchain based ecosystems (Web3) introduce distinct compliance challenges: pseudonymity, high transaction velocity, and jurisdictional fragmentation. Fraud rates in NFTs, for example, range between 2% and 96% depending on market design (Niu et al., 2024; von Wachter et al., 2021), far exceeding those in traditional finance. These environments demand compliance systems that are scalable, explainable, and adaptive to novel risks.

*Agentic AI and Autonomous Orchestration*: Recent work in IS and AI research positions agentic AI as systems where autonomous agents orchestrate workflows and embed goal-directed reasoning (Acharya et al., 2025; Mukherjee & Chang, 2025; Sapkota et al., 2025). A recent literature review maps the evolution of Generative AI (GenAI) agents in financial applications, including fraud detection, compliance monitoring, and advisory (Joshi, 2025). While highlighting growing industry interest in using LLM-based agents, this author also identifies unresolved concerns around traceability, workflow integration, and hallucination risks – underscoring the need for structured frameworks to embed GenAI into compliance-sensitive functions.

Despite this interest, applications of agentic AI to financial crime compliance remain scarce. Our scoping review of SCOPUS and the AIS eLibrary revealed few directly relevant studies: while explainable transaction monitoring has been addressed (Gerlings & Constantiou, 2023) and agentic systems have been examined in domains such as risk governance (Okpala et al., 2025), no prior research demonstrates how agentic AI can coordinate end-to-end FCC processes such as onboarding, monitoring, investigation, and reporting.

*Problematisation and Research Gap*: Collectively, prior research highlights three enduring tensions. First, there is a compliance effectiveness gap: escalating costs and volumes of suspicious activity reports do not translate into proportional enforcement outcomes (Pol, 2020). Second, there is a design knowledge gap: while compliance-by-design and XAI principles are theorized, few actionable design principles exist for implementing AI systems in FCC. Third, there is a contextual gap – research on RegTech and AI-enabled compliance largely addresses traditional financial institutions, while digitally native platforms such as NFT marketplaces remain underexplored.

Against this backdrop, our work contributes to the IS discourse by (1) defining and operationalizing agentic compliance as a design paradigm for FCC; (2) developing a reference architecture and prototype system that embed compliance constraints, explainability, and structured handovers directly into agentic workflows; and (3) deriving design principles that extend existing IS theory on compliance-by-design and sociotechnical governance. In doing so, this research responds to calls in IS design science (Gregor & Hevner, 2013; Sein et al., 2011) and RegTech scholarship (Gozman et al., 2018) to move beyond descriptive applications and develop transferable design knowledge for high-stakes, regulated domains.

## 3. Method

We adopt an Action Design Research (ADR) approach (Sein et al., 2011) to guide the development and deployment of an agentic AI system for FCC. ADR is well-suited as it integrates problem formulation, artifact building, and organizational learning in iterative cycles, allowing the system to be shaped jointly by researchers, practitioners, and regulators.

ADR is particularly suitable for this study because FCC represents a complex, high-stakes problem domain where solutions must be both practically viable and theoretically grounded. Following (Sein et al., 2011), ADR enables the joint evolution of problem framing, artifact development, and organizational learning with stakeholders, which is critical in regulated contexts where compliance obligations are dynamic and contested. Our design choices also reflect this orientation: descriptive analytics and rule-based logic were prioritized to ensure auditability and regulatory trust, while LLM components were introduced in a controlled manner to enhance flexibility without undermining explainability. Workflow orchestration further supported compliance-by-design by embedding guardrails, structured handovers, and full logging directly into system processes. Together, these decisions illustrate how the ADR process was guided not only by technical feasibility but also by the institutional and regulatory demands of the FCC domain.

The project was conducted in collaboration with a fintech startup applying for licensing under the EU's MiCA regulation. Participants included system developers, compliance officers, and regulatory stakeholders. The ADR process followed four rapid build–intervene–evaluate loops over an eight-week period, focusing on automating onboarding, transaction monitoring, alert triage, case investigation, and reporting. This ensured both empirical grounding and progressive refinement of design principles.

Our design process evolved alongside ongoing dialogue about the role of human compliance officers and the transition from the traditional 'three lines of defense' model to agentic orchestration with structured escalation paths (Axelsen et al., 2023).

To evaluate the artifact in a realistic setting, we constructed a domain-specific dataset from OpenSea, the leading NFT marketplace. As of May 2025, this included 816.227 transactions across 859 gaming-related collections. Transactions were risk-scored using a rule-based and prescriptive analytics framework informed by FATF guidance, AML typologies, and Web3-specific crime patterns (e.g., wash trading, obfuscation). This produced 3.055.724 alerts, aggregated at transaction and wallet levels to simulate real-world FCC workloads. The higher alert volume reflects trade duality (buyer and seller), frequent detection of new wallets, and transactions triggering multiple alert types. Viewed by transaction only, this equates to 4,5 pct suspicious transaction activity in the gaming collections vs 21 pct in the overall OpenSea dataset. Figure 2 presents the distribution of triggered alerts.

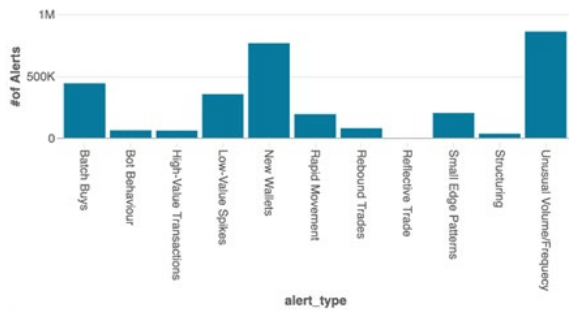

**Figure 2. Distribution of alerts.**

We implemented the artifact using OpenAI's Agent SDK and n8n, a workflow automation tool capable of handling AI agents, to combine LLM reasoning with deterministic tasks, embedding guardrails, escalation triggers, and traceability, while leaving the architecture extensible to predictive models. Together, the methods provided a structured, real-world testbed for developing and refining design principles for agentic AI in FCC.

The prototype prioritized descriptive analytics and rule-based logic to automate core FCC workflows, with architecture designed to accommodate future predictive models through a controlled enrichment process. OpenAI's Agent SDK was selected for its ability to combine LLM-driven reasoning with deterministic, rule-based tasks in a way that is auditable and modular. Execution was orchestrated in n8n, chosen over alternatives for its extensibility and support for a Model Context Protocol (MCP), enabling dynamic model routing and seamless integration with compliance-relevant systems such as KYC, blockchain analytics, and sanctions screening. Together, this stack provides a secure, low-latency, and extensible foundation for intelligent compliance automation in regulated environments.

## 4. Requirements

The requirements for the prototype were derived from European regulatory frameworks, in particular the 5th and 6th Anti-Money Laundering Directives. In addition, the Markets in Crypto-Assets (MiCA) regulation and the forthcoming EU AI Act introduce new obligations relating to transparency, explainability, and oversight of AI systems. Taken together, these frameworks set out a baseline for Financial Crime Compliance (FCC) while also shaping expectations for trustworthy AI. The system was set to cover 6 of the 8 required competencies and activities in our scope for FCC agentic AI, table 1.

| 1 | Know Your Customer (KYC) / Business (KYB) |
|---|---|
| 2 | Transaction Monitoring |
| 3 | Internal Investigation |
| 4 | Escalation and Internal Notification |
| 5 | Reporting Obligations |
| 6 | Recordkeeping |
| 7 | Policies, Training, and Governance |
| 8 | Ongoing Risk Assessment and Program Review |

**Table 1. Scope of the FCC Agentic AI**.

In translating these directives into system design, our ADR process stipulated requirements for auditability, risk-based orchestration, agent–agent collaboration, and human oversight. To meet these, we specified the need for multiple specialized agents (e.g., investigation agent, anomaly-detection agent, reporting agent), each fulfilling a distinct role. This modular architecture ensures deterministic outcomes, fast execution, and structured oversight – without positioning humans as the primary counterpart in every decision. It reflects a shift toward agentic compliance, where workflows are distributed across cooperating agents to maintain regulatory alignment and enable future extensibility with AI models.

## 5. Artifact

Building on (Lohmann, 2013)'s principle of compliance-by-design, we extend compliance logic beyond static workflows into autonomous agent behavior and decision-making rules. Agents are orchestrated around compliance artifacts (e.g., KYC records, alerts, investigation, case management, STRs, reporting, record keeping), embedding regulatory rules structurally into their operation.

Through four ADR iterations, we developed a prototype FCC system capable of automating onboarding, transaction monitoring, alert triage, case investigation, and reporting and record keeping. Key design insights included: (i) modularizing context beats simply scaling models; (ii) deterministic checks belong upstream in rule logic; and (iii) observability through logging and instrumentation provides the fastest lever for design improvement. These insights informed a layered architecture where deterministic controls anchor workflows and LLM agents provide explainability and augmentation. Figure 3 illustrates the sequencing of FCC tasks within the agentic framework.

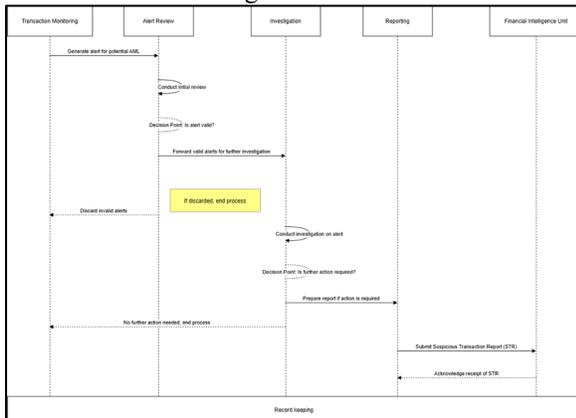

**Figure 3. FCC sequencing diagram.**

To operationalize agentic compliance, three design principles were instantiated: (1) Embedded guardrails and handovers – consistent with compliance-by-design, guardrails constrain agent actions via permissions, rules, and thresholds, while structured handovers (agent–agent or agent–human) manage exceptions and escalation; (2) explainability and auditability – following SR 11-07 guidance (FED, 2011), all agent decisions are logged with rationales, ensuring transparency, reproducibility, and accountability. Novel elements include a semantic cache and reinforcement cache, enabling traceability while iteratively balancing false positives and missed alerts; (3) extensibility and risk-based orchestration, extending modular design principles (Gregor & Hevner, 2013), agents coordinate through transaction- and wallet-level risk scores that trigger enhanced due diligence or reporting. Predictive models can be integrated through a controlled enrichment process, informing but not dictating compliance actions.

These principles combine established IS constructs (compliance-by-design, traceability, audit logging) with novel extensions (agent–agent orchestration, semantic caching for explainability, reinforcement cache for performance optimization). Figures 4–6 illustrate the layered architecture, agent orchestration, and learning loop, where Figure 4 shows the AI agent framework architecture implementation in n8n UI. In the center of Figure 4 the FCC agent orchestrates the agentic AI, by receiving data triggers from the Postgres database with required data transformations.

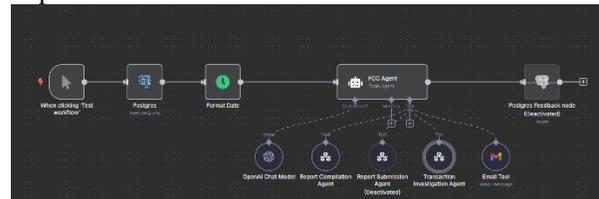

**Figure 4. FCC Agentic AI architecture.**

The postgres alerts are analyzed via the Investigation AI agent's understanding of risk scoring via algorithms, semantic cache or reinforcement cache, as illustrated in Figure 5.

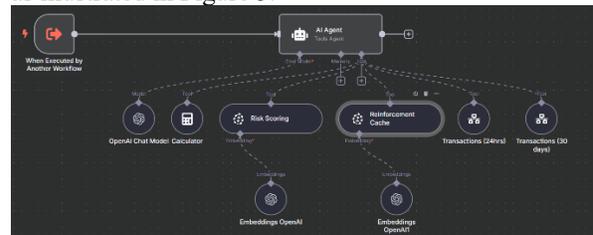

**Figure 5. Investigation AI agent architecture.**

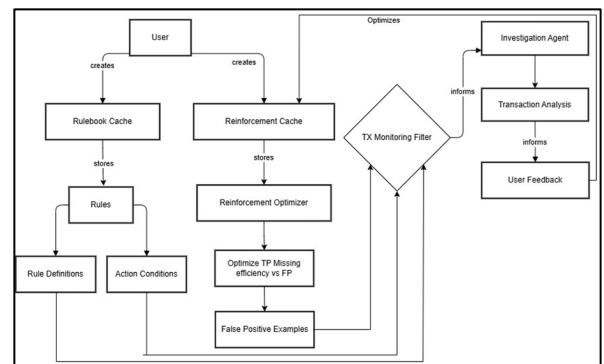

**Figure 6. Learning loop and semantic cache.**

Figure 6 shows a diagram of associated learning loop, the reinforcement cache, where the transaction analysis provides feedback to a semantic cache with a mix of rules and algorithms, updated through a learning optimizer balancing True Positive missing vs False Positive alerts.

## 5.1. Results

The prototype generated structured case files that integrated transactional risk scores, behavioral indicators (e.g., repetitive trades, new wallet activity), and regulatory checks into outputs suitable for SAR/STR reporting as shown in Figure 7 output.

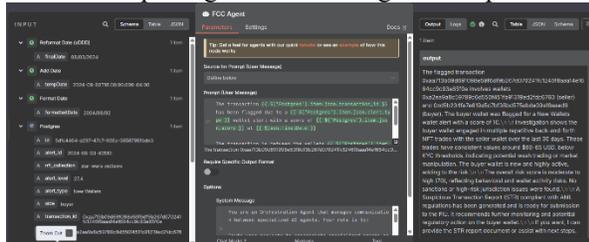

**Figure 7. FCC agent input-throughput-output.**

'*The flagged transaction 0xaa713b09d691098e59f-6df9b267d070241fc1246f8aaa14e1664cc9c93e85f0e involves wallets 0xa2ea9a8c59789c6d550f451fb91-319ed2fdc6760 (seller) and 0xd5b234fa7e619a-5c7bf38bd575abde09a18eaed6 (buyer). The buyer wallet was flagged for a New Wallets wallet alert with a score of 10. Investigation shows the buyer wallet engaged in multiple repetitive back-and-forth NFT trades with the seller wallet over the last 30 days. These trades have consistent values around $60-65 USD, below KYC thresholds, indicating potential wash trading or market manipulation. The buyer wallet is new and highly active, adding to the risk. The overall risk score is moderate to high (70), reflecting behavioral and wallet activity risks. No sanctions or high-risk jurisdiction issues were found. A Suspicious Transaction Report (STR) compliant with AML regulations has been generated and is ready for submission to the FIU. It recommends further monitoring and potential regulatory action on the buyer wallet.*'

Beyond case assembly, the system also demonstrated the ability to trigger alerts, initiate escalations, and maintain auditable records across the FCC process. Each workflow execution left a traceable log of agent actions, handovers, and rationales, ensuring alignment with AML directive requirements and providing a governance layer consistent with supervisory expectations. This layered capability shows that agentic orchestration can cover not only case generation but also upstream detection and downstream reporting, creating an end-to-end compliance pipeline. Although predictive modeling is not yet embedded, the architecture supports future integration under conditions that preserve explainability and regulatory alignment.

## 6. Evaluation

The artifact was evaluated through a formative, naturalistic approach consistent with ADR (Sein et al., 2011), emphasizing how agentic AI enables explainable orchestration of FCC workflows under regulatory constraints. Rather than focusing on the accuracy of individual detection rules, the evaluation targeted the system's architecture and ability to ensure compliance-by-design through traceable, autonomous agents.

Evalution was formative as regulatory assessment was ongoing without formal acceptance. Structured walkthroughs using data from OpenSea and real-world deployments assessed key use cases: onboarding checks, transaction monitoring, and escalation logic. Each agent operated within a clearly defined mandate, triggering discrete, auditable actions with full logging. This task-based decomposition supports traceability, oversight, and modular transparency, core tenets of explainable AI, even in the absence of formal model-level explainability (e.g., LIME, SHAP).

Evaluation applied principles from the Co-12 XAI framework (Nauta et al., 2023) focusing on correctness, completeness, coherence, and compactness of the agents' decisions. Output explanations were assessed against the Danish FIU's good notification template using scenario-based reviews by compliance stakeholders. These confirmed that the system provided meaningful, actionable STRs with particular value in the translation of alerts to cases that are context-relevant and procedurally justified. Prescriptive analytics modelling and rules based logic was covered by the evaluated prototype, and the platform was designed for future integration of predicitive model-based enrichment. The agentic architecture ensures that as models are introduced, explanations remain embedded within traceable agent workflows. Evaluation of predictive components will prioritize not only precision but also interpretability, latency, and cost-efficiency, aligned with US FED SR 11-7 model governance standards (FED, 2011).

For the prototyping we tested the agentic system's general ability to provide justification of cases when translating alerts to case. Particular attention should be given to the precision–recall trade-off, since false positives inflate costs while false negatives pose regulatory risks. In our prototype, applying a general risk-tolerance threshold produced few escalations, highlighting that threshold calibration will be critical and will likely remain an ongoing discussion with regulators. Models that fail to meet baseline compliance thresholds, whether due to opacity, inconsistency, or inference lag, should be downgraded or excluded from the semantic cache. A noted trade-off is that reliance on a central coordinating agent under the Model Context Protocol (MCP) introduces tight coupling and potential bottlenecks (Yang et al., 2025). While the current implementation does not yet integrate full MCP coordination, the system is architected to support it. This readiness reflects the demands of regulated FCC environments, where centralized control is often necessary for accountability and human oversight. The artifact demonstrates how agentic AI can operationalize regulatory compliance, explainability, and workflow automation in parallel, laying the groundwork for a risk-aware FCC system that is auditable, institutionally aligned, and cost-efficient..

Overall, evaluation aligns with the expectations for relevance and rigor in Action Design Research (Mullarkey & Hevner, 2019), demonstrating that the artifact provides a viable pathway toward agentic, explainable, and regulation-aligned FCC systems, even in early stages of predictive integration. This combined approach of real-world grounding, LLM-assisted labeling, and synthetic replication provides a robust basis for initial system design and iterative testing of risk analytics and workflow orchestration within the FCC context.

### 6.1 Cost Implications

While the system is not yet deployed at scale, early-stage estimates suggest meaningful efficiency gains. Industry benchmarks report that the average suspicious transaction report (STR) requires approximately 1,98 hours of analyst time, with complex cases taking over 20 hours (RegTech Consulting, 2020). By contrast, the prototype automates triage, case assembly, and report drafting in under one minute per case, including regulatory handover formatting. Even with human oversight, this implies a significant reduction in time per STR, potentially reducing compliance effort by more than 98% across the reporting lifecycle.

To illustrate the relevance for digitally native industries, consider a hypothetical Web3 gaming studio with 20 employees and a user base of 100,000 gamers, each making approximately 100 transactions per year. At a 4,5% fraud suspicion rate, this could result in 450.000 alerts annually. With a traditional manual process (2 hours per alert), this would require over 900.000 analyst hours, equivalent to more than 480 full-time compliance staff. Such a burden would be prohibitive for small firms.

In contrast, deploying the agentic AI system with OpenAI's GPT-4.1-mini at current pricing (approximately US$1 per 1,666 API calls) yields a projected inference cost of just US$600 per year, possibly less for local models. Even if a small compliance team of 6–8 FTEs were retained for oversight, the cost structure becomes more manageable, especially if offset by the higher margins typical in Web3 gaming compared to traditional models. These estimates remain illustrative and subject to change based on integration maturity, regulatory acceptance, and organizational readiness. They will require validation through field deployment and regulatory acceptance. Nonetheless, they suggest that agentic automation could materially alter the economics of compliance, particularly for smaller institutions facing disproportionate regulatory burdens (LexisNexis, 2024).

## 7. Discussion

This research set out to design and implement an agentic AI framework to improve the efficiency, transparency, and regulatory alignment of financial crime compliance (FCC) processes particularly around the translation of risk alerts to case handling and reporting, responding to the following research question: *How can agentic AI systems be designed to support scalable, explainable, and regulation-aligned FCC in digitally native financial platforms?*

The resulting artifact automates key compliance tasks – such as onboarding, transaction monitoring, escalation, and reporting – within a structured workflow environment. By embedding compliance requirements directly into agent behaviors and orchestrating actions through explainable and traceable logic, the artifact demonstrates utility in supporting both operational and regulatory needs. The system's architecture supports modular expansion and ongoing monitoring, making it adaptable to evolving compliance expectations.

The artifact contributes to IS design theory by surfacing a set of design principles for agentic AI in high-risk, regulated domains. These include (i) embedding compliance constraints as agentic guardrails, (ii) enabling structured handovers for human oversight, (iii) leveraging explainable logic in decision chains and continuously evaluating those, and (iv) integrating semantic model selection for future predictive extensions. These principles are applicable beyond FCC and offer a reusable foundation for designing autonomous systems in other regulatory or mission-critical contexts.

These principles extend existing IS knowledge by adapting established constructs such as compliance-by-design (Lohmann, 2013), and auditability in model governance (FED, 2011), while introducing novel mechanisms including agent–agent orchestration, semantic model routing, and reinforcement caches for performance optimization.

The implementation also reveals how agentic AI reshapes work practices in compliance. Rather than fully replacing human actors, the system redistributes roles toward oversight, exception handling, and interpretive review. Compliance officers can become curators of edge cases and stewards of model governance rather than manual processors, and the system itself prevents conflict-of-interest, perhaps allowing a leaner organization of the compliance function in regulated entities. This reinforces themes in IS literature on sociotechnical system evolution, highlighting how AI augments rather than eliminates human judgment. The agentic framework enables a form of hybrid intelligence where automation and accountability co-exist within formal organizational processes.

Compared to existing RegTech solutions such as, for example, ComplyAdvantage (Complyadvantage, 2025), Sumsub (Sumsub, 2025), or ShuftiPro (ShuftiPro, 2025), which increasingly integrate GenAI for fraud detection, sanctions screening, or onboarding automation, our approach differs first and foremost by holistic orchestration of end-to-end FCC processes, including case management, across multiple agents with embedded explainability and guardrails. While commercial tools tend to be focused on specific tasks such as identity verification or alert filtering, the artifact presented in this case demonstrates how agentic compliance can integrate risk scoring, orchestration, and regulatory reporting into a coherent, regulation-aligned framework.

From a policy perspective, the artifact demonstrates how AI systems can be designed to comply with both industry-specific expectations for model risk governance, and emerging cross-sectoral regulation, such as the EU AI Act. The integration of guardrails, explainability layers, and human handovers aligns well with the AI Act's obligations for high-risk systems, including transparency, human oversight, and post-deployment monitoring. The work offers a possible model for how digital-first compliance systems can meet regulatory standards not through post hoc audits but through embedded and enforceable logic. The implementation also invites a reassessment of the prevailing 'three lines of defense' model, where siloed roles occupied by humans reinforce process-heavy compliance with limited risk reduction and high costs.

Our findings confirm that digital systems can collapse unnecessary handoffs and make compliance more responsive, evidence-based, and integrated. This is especially relevant given the tension between expansive reporting requirements, and imbalances in compliance capacity or insufficient resources for effective follow-up. Hence, the current compliance paradigm may have become too risk-averse, prioritizing procedural coverage over actual regulatory outcomes. A digitally native approach could shift the emphasis from box-ticking toward intelligent, risk-weighted enforcement, improving not only efficiency but also the socio-economic returns of regulation.

While the system shows promise, it is limited in several respects. The current implementation lacks predictive depth, relying primarily on descriptive analytics and rule-based logic and still a level of fragmentation of the FCC process. The evaluation is artifact-centered rather than field-tested in a larger operational setting, and human/regulatory acceptance, organizational change dynamics, and long-term trust mechanisms are not yet explored.

Future work could deepen the predictive capabilities through controlled semantic model sourcing, extend the governance layer to handle more adaptive AI components, and test the approach in multi-jurisdictional regulatory environments to evaluate scalability and transferability.

As a further reflection, this study demonstrates how low-code tools and agentic AI architectures can reimagine financial crime compliance as a digitally native process. It contributes to IS research by showing how design science methods can automate sensitive, high-stakes domains without sacrificing transparency, auditability, or regulatory alignment. The findings support a broader view of AI as a governance technology, one that demands not only technical sophistication but also attention to institutional roles, decision rights, and procedural accountability.

While this work focuses on FCC, the design principles of agentic compliance – structured handovers, explainability layers, and eventually semantic model routing via MCP – may generalize to other high-stakes domains such as healthcare audits, environmental reporting, or supply chain compliance. These fields share characteristics like regulatory pressure, complex data landscapes, and a need for traceable decision-making.

Future research could adapt the agentic orchestration model to such settings, testing its capacity to embed institutional logic into autonomous systems more broadly. This extends IS literature on digital governance and design theory by proposing agentic orchestration as a reusable pattern in contexts where automation, compliance, and institutional accountability must coexist.

Finally, the findings highlight a broader systemic issue in FCC: rising volumes of SAR/STR reporting coupled with limited regulatory outcomes. The AML regime has been characterized as a form of "third-party policing," in which compliance responsibilities are outsourced to private actors with often limited effectiveness (Nazzari & Reuter, 2025). Agentic compliance may offer a way to rebalance this arrangement by linking reporting effort more directly to risk severity and automation quality.

In jurisdictions where high reporting volumes contrast with limited follow-up capacity and poor system integration, this approach could reduce inefficiencies while improving both cost-effectiveness and regulatory utility.

## 8. Conclusion

This study addresses inefficiencies and escalating costs in financial crime compliance (FCC) by designing an agentic AI framework that automates key compliance workflows while holistically embedding regulatory alignment, traceability, and explainability. Rather than focusing on optimizing detection algorithms, the contribution lies in orchestrating how tasks such as onboarding, transaction monitoring, and suspicious activity reporting can be delegated to autonomous agents under structured governance.

The artifact was developed using a DSR approach embedded within an ADR-setting, allowing for iterative refinement through real-world engagement with a fintech operating in a regulated domain. By leveraging descriptive analytics over a curated dataset of game-related NFT transactions from OpenSea, combined with synthetic data generation, we evaluated the framework's ability to support agent-led orchestration under compliance-by-design principles. As with any formative ADR effort, evaluation remains early-stage and regulatory review is ongoing, but the prototype provides a foundation for further testing and institutional engagement.

Theoretically, this work contributes to IS research on AI-human collaboration, regulatory technologies, and agent-based systems by demonstrating how autonomous agents can be operationalized within institutional constraints. The inclusion of a Model Context Protocol (MCP) points toward a potential advance in AI governance mechanisms, complementing established constructs, such as compliance-by-design and auditability. While broader testing of local model routing in this case is still pending, MCP illustrates how compliance agents could reason about model selection based on task-level factors such as explainability, jurisdiction, or cost.

While the system currently operates with descriptive logic and synthetic data, future development will focus on integrating predictive components, including LLMs and structured anomaly detection models. Despite progress in machine learning, financial crime compliance systems based on LLMs often lack integration between LLMs and structured analytics, rely on post-transaction detection, and offer limited explainability (Korkanti, 2024). So, apart from fine-tuning and fitting LLMs to predictively detect and then integrate this with the structured analytics in place, evaluating these using rigorous performance and explanation metrics is another challenge. In particular, future work would look to apply the Co-12 framework (Nauta et al., 2023) or similar frameworks meeting upcoming EU regulatory technical standards to assess explanation quality across dimensions such as correctness, coherence, and completeness, ensuring that agentic decisions are accurate, interpretable and auditable.

While our prototype demonstrates how agentic compliance can embed regulatory logic directly into autonomous workflows, we recognize that full replacement of human oversight remains unrealistic at this point. As highlighted in recent industry blogs, accuracy degradation across multi-step verification chains and unresolved accountability for regulatory breaches mean that human expertise will remain central (Malyutin, 2025). Our findings align with this perspective; agentic AI is best positioned not as a substitute for compliance officers but as an orchestrator that reallocates human effort toward high-value investigative and governance tasks.

Future research should therefore move beyond the binary of replacement versus augmentation, examining how agent–agent orchestration and human oversight can be balanced to optimize both efficiency and accountability. This requires (i) evaluating multi-agent systems across longer compliance chains, where error compounding is a risk; (ii) testing handover mechanisms under regulatory review; and (iii) assessing how such systems reshape the compliance workforce over time. By engaging with both academic design theory and evolving industry practice, IS research can help steer the development of agentic AI toward trustworthy, regulation-aligned adoption in FCC and related high-stakes domains.

As regulatory expectations rise, including under the EU AI Act, this research offers a forward-looking model for designing AI systems that balance automation and accountability. It contributes both a practical instantiation and a set of transferable design principles for building compliance-aligned, explainable, and agentic information systems in high-stakes domains, while recognizing that evaluation remains formative and subject to regulatory validation.